# CallCenterEN: 91706 Real-World English Call Center Transcripts Dataset with PII Redaction

## Authors

Ha Dao [1]     Gaurav Chawla [2]     Raghu Banda [3]     Caleb DeLeeuw [4]

[1] AIxBlock, USA

[2] [4] Independent Researchers, USA

[3] INSEAD Business School

## Abstract

We introduce CallCenterEN, a large-scale (91,706 conversations, corresponding to 10448 audio hours), real-world English call center transcript dataset designed to support research and development in customer support and sales AI systems. This is the largest release to-date of open source call center transcript data of this kind. The dataset includes inbound and outbound calls between agents and customers, with accents from India, the Philippines and the United States. The dataset includes high-quality, PII-redacted human-readable transcriptions. All personally identifiable information (PII) has been rigorously removed to ensure compliance with global data protection laws. The audio is not included in the public release due to biometric privacy concerns. Given the scarcity of publicly available real-world call center datasets, CallCenterEN fills a critical gap in the landscape of available ASR corpora, and is released under a CC BY-NC 4.0 license for non-commercial research use.

## 1. Introduction

There is a growing need for high-quality, real-world datasets for training AI models in specialized domains like customer service and sales [1, 2, 3]. AI systems such as Large Language Models (LLMs) benefit significantly from training on specialized domain-specific data [4]. However, call center audio data is rarely available due to privacy, regulatory, corporate policies, and logistical challenges.

CallCenterEN aims to address this by releasing a large set of transcriptions from real call center interactions, collected via collaborations with multiple international Business Process Outsourcing (BPO) centers. These conversations reflect a wide variety of realistic commercial support topics, accent profiles, and dialog structures.

The transcriptions represent calls between agents and customers, where agents speak with a range of Indian, Filipino and American accents, and customers are primarily from the United States. 0.1% of the dataset was reviewed by human experts. Both inbound and outbound call flows are included.

## 2. Motivation, Authorship, and Contributions

This dataset was created and published by researchers committed to advancing open AI. Our motivation stems from the lack of high-quality, large-scale, and real-world call center data that can be used for training and evaluating conversational AI models, particularly in the customer service and sales domains. By releasing CallCenterEN, we aim to provide the global research community with a reliable dataset for training domain-specific language models while respecting the ethical and legal boundaries of data sharing.

## 3. Related Work

Existing open speech datasets have laid critical groundwork for Automatic Speech Recognition (ASR) and speech understanding. LibriSpeech [5] provides audiobooks with aligned transcripts and remains a foundational dataset. Common Voice [6] by Mozilla introduces multilingual coverage with crowdsourced contributions but lacks contextual flow and task-oriented dialogue. The Fisher Corpus [7] includes telephone conversations but is outdated and non-commercial in nature.

While datasets like the **Wall Street Journal (WSJ)** corpus and **TIMIT** are historically significant, they feature clean, read speech in studio-like conditions—lacking the spontaneous and noisy nature of real-world calls [8, 9]. The **Switchboard** and **Fisher** corpora, although closer to telephone conversations, are outdated and primarily conversational without specific task or domain orientation. In our release we have not put out the audio recordings but remain open to collaborations using this data, thus these future collaborations could offer comparable speech audio data to these studies: The Fisher Corpus and SWITCHBOARD: Telephone speech corpus for research and development [10, 11].

In contrast, **CallCenterEN** includes modern, business-specific contexts with inbound and outbound structures, suitable for customer support modeling. CallCenterEN differentiates itself by focusing on commercial dialog with real-world accents and structured support scenarios, filling the gap where prior resources fall short. Furthermore, it complements efforts such as the Switchboard corpus and recent dialogue datasets like MultiWOZ by offering more natural, unscripted, and transactional conversation data. There are a variety of call center datasets available on Hugging Face, which are generally much smaller (nearly always fewer than 1000 observations and often fewer than 100 observations) and less complete than we set out to deliver here [12] [13].

## 4. Methodology

The CallCenterEN dataset was sourced through partnerships with approximately 9-10 BPO centers. These partnerships were originally established for commercial data collection purposes, with the dataset subsequently being released for academic research. The selection of BPO centers was driven by domain-specific requirements from previous commercial clients.

The dataset exhibits a geographic distribution reflecting real-world call center operations. For inbound calls, both agents and customers are US-based. For outbound calls, agents are located in India and the Philippines while customers are US-based. Audio recordings were captured around 2020, with dataset compilation completed by 2025. A random sampling methodology was employed across all participating BPO centers to ensure representative coverage.

All audio recordings were obtained in raw, uncompressed format and processed using AssemblyAI's premium commercial ASR service. The transcription process followed the guidelines detailed in Section 8, essentially basic systematic standardization.

The ASR processing generated word-level timestamps and individual confidence scores for each word. Overall confidence scores range from 86–98% across the dataset. A subset representing 0.1% of the total dataset was manually reviewed by a team of experienced transcribers for quality validation. This manual evaluation calculated Word Error Rate (WER), insertions, deletions, and substitutions as shown in the validation table, yielding an average WER of 3.87%.

Both audio and corresponding transcripts underwent PII redaction based on the categories listed in Section 8. The redaction process combined automated detection with manual review to ensure comprehensive removal of personally identifiable information, ensuring compliance with privacy regulations including CCPA and DPDP 2023.

# 5. Dataset Overview

*Table 1: Dataset overview*

| Category | Value |
|---|---|
| Total transcripts | 91,706 |
| Total hours of corresponding audio | 10,448 hours (before removal due to regulatory and biometric privacy concerns) |
| Call types | Inbound, Outbound |
| Language | English |
| Accents | Indian, Filipino, American |
| Annotation method | Paid ASR (AssemblyAI) |
| File format | JSON |
| Confidence score | 86–98% |
| PII redacted | Yes (see Section 8) |
| License | CC BY-NC 4.0 |
| Intended use | <ul><li>Detailed intent detection and classification,</li><li>Domain-specific raw text training corpus for neural network training, especially LLMs,</li><li>Fine-tuning datasets for large language models,</li><li>Creating a benchmark for AI agents to reach the quality level of human agents based on successful conversations,</li><li>For outbound conversations, by categorizing the outputs (successful or not) of the conversation, you can create a benchmark for sales agents. And same for inbound calls, then you can create a benchmark for customer support agents,</li><li>Dialog summarization,</li><li>Sentiment analysis,</li></ul> |

| | - Placeholder-based NER and de-identification model evaluation,<br>- Studying entity context patterns in real-world support conversations,<br>- LLM fine-tuning for support agents,<br>- Generate corresponding synthetic call center data audio. |
|---|---|
| Link to download dataset | https://huggingface.co/datasets/AIxBlock/91706-real-world-call-center-scripts-english |
| WER on test subsets | 3.87% |
| Average accuracy level | 96.131% |

**Domain (topic) distribution**

*Table 2: Domain (topic) distribution*

| Domain/Folder name | Number of conversations |
|---|---|
| Auto_insurance_customer_service_inbound (USA-USA) | 1,749 |
| Automotive_and_healthcare_insurance_inbound (USA-USA) | 1,793 |
| Automotive_inbound (USA-USA) | 6,045 |
| Customer_service_general_inbound (USA-USA) | 1,217 |
| Home_service_inbound (USA-USA) & telecom_outbound (Indian- USA) | 3,239 |
| Home_service_inbound (USA- USA) | 11,407 |
| Insurance_outbound (Indian - USA) | 4,005 |
| Medical_equipment_outbound (Filipino) | 738 |
| Medicare_inbound (USA- USA) | 61,513 |

As shown in Figure 1 below, the majority of domains fall within the medical sector, with the largest constituent being Medicare inbound calls, making this dataset a highly valuable resource for training models in the medical field.

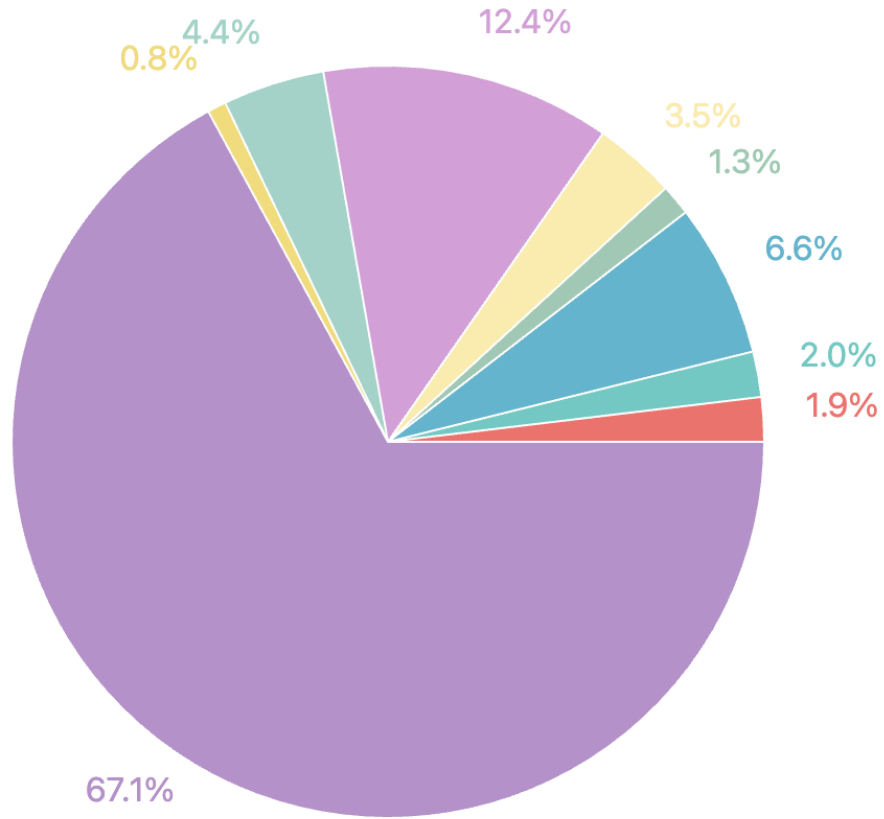

(a) Pie chart showing proportional distribution

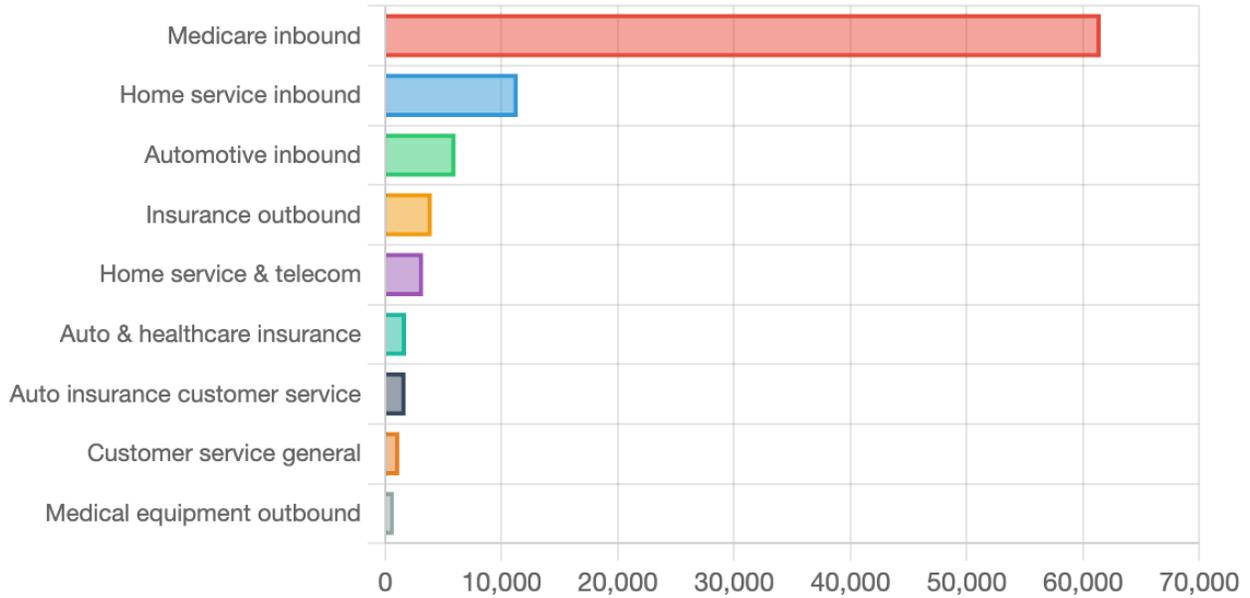

*(b) Horizontal bar chart showing conversation counts*

*Figure 1: Domain distribution in CallCenterEN dataset. (a) Proportional distribution across nine domains. (b) Absolute conversation counts per domain. Medicare inbound calls dominate with 67.1% of total conversations.*

**Inbound/outbound distribution**

*Table 3: Inbound/outbound distribution*

| Call type | Count | Percentage |
| --- | --- | --- |
| Inbound | 83,724 | 91.3% |
| Outbound | 7,982 | 8.7% |

Figure 2 shows that the majority of calls are inbound, making this dataset a valuable resource for training models on customer support interactions. Inbound calls typically involve customers reaching out to agents for assistance, whereas outbound calls are initiated by agents, often for cold-calling or upselling purposes.

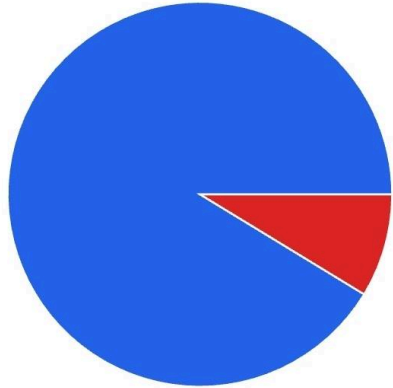

*Figure 2: Inbound and Outbound Distribution*

Each JSON transcript begins with the complete PII-redacted conversation, followed by metadata such as the overall confidence score (reflecting ASR accuracy) and the duration of the audio in seconds. The transcript then provides word-level timestamps, where each word is paired with its individual confidence score. This structure enables detailed temporal analysis and fine-grained error tracking for speech recognition or dialogue modeling experiments.

**6. PII Redaction & Compliance**

All transcripts have been redacted to remove personally identifiable information (PII) using a combination of automated entity detection and manual review. Below is the list of PII categories that were reviewed and removed from the transcripts when detected:

- Personal Identifiers: person_name, person_age, date_of_birth,

- phone_number, email_address, physical_attribute, gender_sexuality, marital_status, nationality, religion, political_affiliation, zodiac_sign,
- Financial Information: account_number, banking_information, credit_card_number, credit_card_expiration, credit_card_cvv, money_amount,
- Government/Legal Identifiers: drivers_license, passport_number, us_social_security_number, vehicle_id,
- Medical Information: blood_type, healthcare_number, medical_condition, medical_process, drug, injury,
- Technical/System Data: ip_address, url, username, password, filename,
- Temporal Information: date, date_interval, time, duration,
- Location/Organization: location, organization, occupation,
- Other: event, language, number_sequence, statistics.

These redactions were applied to both structured and unstructured text spans to ensure full anonymization. The dataset complies with relevant privacy regulations in the United States (e.g., CCPA) and India (DPDP 2023).

**7. Data Acquisition and Processing**

The dataset was sourced through partnerships with multiple BPO centers operating across various domains. All recordings were downloaded in raw, uncompressed audio format to preserve acoustic fidelity.

Calls were then categorized by domain, accent (Indian, American, Filipino), and topic (inbound vs outbound).

After domain tagging, we used AssemblyAI's advanced paid ASR engine to transcribe the recordings. Then both audio and its corresponding transcripts had PII redacted based on the predefined list of PII info.

Each transcript includes word-level timestamps and ASR confidence scores.

A post-processing step was then applied to clean the data, manually QA a test subset to calculate WER.

Below is a sample of a small test set human QA result (in total, we manually tested 0.1% of the total dataset).

*Table 4: A small example of how we conducted human QA*

| File Number | ASR Word Count | QA Word Count | Insertions | Deletions | Substitutions | Total Errors | Word Error Rate (%) | Accuracy level |
|---|---|---|---|---|---|---|---|---|
| 1 | 1,752 | 1,768 | 26 | 10 | 19 | 55 | 3.11% | 96.89% |
| 2 | 2,559 | 2,630 | 86 | 15 | 36 | 137 | 5.21% | 94.79% |
| 3 | 1,363 | 1,416 | 59 | 6 | 21 | 86 | 6.07% | 93.93% |
| 4 | 2,673 | 2,761 | 95 | 7 | 45 | 147 | 5.32% | 94.68% |
| 5 | 1,098 | 1,125 | 29 | 2 | 12 | 43 | 3.82% | 96.18% |
| 6 | 906 | 957 | 51 | 0 | 16 | 67 | 7.00% | 93.00% |
| 7 | 568 | 593 | 25 | 0 | 14 | 39 | 6.58% | 93.42% |
| 8 | 522 | 548 | 27 | 1 | 2 | 30 | 5.47% | 94.53% |
| 9 | 369 | 369 | 0 | 0 | 0 | 0 | 0.00% | 100.00% |
| 10 | 524 | 535 | 11 | 0 | 5 | 16 | 2.99% | 97.01% |
| 11 | 1,713 | 1,733 | 26 | 6 | 16 | 48 | 2.77% | 97.23% |

| | | | | | | | |
|---|---|---|---|---|---|---|---|
| 12 | 1,413 | 1,421 | 15 | 7 | 24 | 46 | 3.24% | 96.76% |
| 13 | 1,485 | 1,570 | 88 | 3 | 28 | 119 | 7.58% | 92.42% |
| 14 | 635 | 638 | 4 | 1 | 14 | 19 | 2.98% | 97.02% |
| 15 | 1,177 | 1,183 | 8 | 2 | 33 | 43 | 3.63% | 96.37% |
| 16 | 2,988 | 2,979 | 20 | 29 | 72 | 121 | 4.06% | 95.94% |
| 17 | 1,198 | 1,195 | 25 | 28 | 28 | 81 | 6.78% | 93.22% |
| 18 | 903 | 899 | 4 | 8 | 7 | 19 | 2.11% | 97.89% |
| 19 | 2,322 | 2,338 | 30 | 14 | 50 | 94 | 4.02% | 95.98% |
| 20 | 485 | 495 | 12 | 2 | 13 | 27 | 5.45% | 94.55% |
| 21 | 1,847 | 1,857 | 21 | 11 | 27 | 59 | 3.18% | 96.82% |
| 22 | 370 | 370 | 1 | 1 | 3 | 5 | 1.35% | 98.65% |
| 23 | 434 | 436 | 4 | 2 | 3 | 9 | 2.06% | 97.94% |
| 24 | 667 | 671 | 10 | 6 | 13 | 29 | 4.32% | 95.68% |

| 25 | 1,020 | 1,020 | 10 | 10 | 33 | 53 | 5.20% | 94.80% |
| 26 | 1,434 | 1,440 | 8 | 2 | 14 | 24 | 1.67% | 98.33% |
| 27 | 1,535 | 1,540 | 6 | 1 | 13 | 20 | 1.30% | 98.70% |
| 28 | 4,383 | 4,388 | 16 | 11 | 40 | 67 | 1.53% | 98.47% |
| 29 | 2,938 | 2,944 | 22 | 16 | 31 | 69 | 2.34% | 97.66% |
| 30 | 582 | 588 | 11 | 5 | 13 | 29 | 4.93% | 95.07% |

Note: Insertions = words incorrectly added by ASR; Deletions = reference words missing from ASR output; Substitutions = words incorrectly transcribed as different words; WER = (Insertions + Deletions + Substitutions) / Total Reference Words.

Given this randomly selected subset test set, we are confident in the overall dataset's accuracy, as the transcription was performed using a state-of-the-art ASR model under a premium service tier.

Accent diversity and domain specificity were preserved during this process to reflect real-world usage patterns.

## 8. Applications

Please note that this is a raw dataset without labels. However, here are some of the intended use cases for this data:

CallCenterEN enables a range of research tasks, including:

- Detailed intent detection and classification,
- Additional rare domain-specific raw text training corpus for neural network training especially LLMs and fine-tuning of LLMs (including the creation of foundational models which may be later

- released open source, which we consider to be a form of research.),
- Recommendation for customer service agents based on satisfied conversation dataset,
- Creating a benchmark for AI agents to reach the quality level of human agents based on successful conversations,
- For outbound conversations, by categorizing the outputs (successful or not) of the conversation, you can create a benchmark for sales agents. And same for inbound calls, then you can create a benchmark for customer support agents,
- Dialog summarization,
- Sentiment analysis,
- Placeholder-based NER and de-identification model evaluation,
- Studying entity context patterns in real-world support conversations,
- LLM fine-tuning for support agents,
- Generate corresponding synthetic call center data audio.
- Classification of call outcomes using the last few sentences of the call as indication of a resolution of the issue or no resolution. The same can be used for outbound cold calls also.

This dataset represents the current state of acceptable human call center task performance behavior, by showing the real-world current state of human call center service in the listed domains (see table 2.) Given the high quality and domain-specific nature of the transcripts, the dataset may help researchers and enterprise R&D departments as a benchmarking resource for conversational AI research. The domain-specific expertise exemplified here is also useful inside and outside those specific domains. As exemplified by Hu et al's observation that "the availability of a specialized microstructure dataset propagates across different areas within finance and into other disciplines", we hope this dataset can be applied to research and benchmarking in a wide variety of domains [14, 4].

## 9. Transcription Guidelines:

The PII categories listed in Section 6 are systematically redacted during transcription.

The following filler words are removed by default:

- "um"
- "uh"
- "hmm"
- "mhm"
- "uh-huh"
- "ah"
- "huh"
- "hm"
- "m"

In addition, all numbers are converted to their numerical form.

## 10. Access, Licensing, and Author Roles

The dataset is publicly available at:

https://huggingface.co/datasets/AIxBlock/91706-real-world-call-center-scripts-english

License: Creative Commons Attribution-NonCommercial 4.0 International (CC BY-NC 4.0).

Use is permitted for academic research and non-commercial model development only. Commercial use, resale, or redistribution is prohibited.

## 11. Legal Disclaimer:

This dataset was collected through authorized partnerships with BPO providers. The publicly released version includes only PII-redacted transcriptions. No audio data is shared to avoid biometric re-identification risks. All personally identifiable information has been removed.

Users of this dataset agree not to attempt re-identification of any individuals or organizations. The dataset is shared for academic and research purposes only, with a strong encouragement of ethical handling.

## 12. Limitations:

Due to the real-world nature of call center recordings, the source audio includes background noise, cross-talk, and other natural artifacts that affect transcription quality. Although we employed one of the most advanced commercial ASR models available at the time, the transcription accuracy may not be perfect in every case. Human-in-the-loop QA was not applied across the full dataset due to budget constraints.

Only 0.1% of the dataset was formally human reviewed, which is below a statistically significant sample size. However, the authors believe this review size is reasonable here because the methods of processing and sourcing were very standardized, and professional reviewer time was limited. Authors have done additional informal review in handling the data during upload to Hugging Face Hub.

Additionally, regulatory restrictions prevent the public release of even redacted audio files. However, we are open to collaboration with trusted institutions that can ensure full legal compliance; if you are interested in accessing the corresponding audio dataset under appropriate safeguards, feel free to contact us at the following email address:

hadao@aixblock.io
raghu.banda@insead.edu

gaurav.chawla@columbia.edu

Evaluation (LREC), 69–71. https://aclanthology.org/L04-1500/

[11] Godfrey, J.J., Holliman, E. (1992). SWITCHBOARD: Telephone speech corpus for research and development. In ICASSP 1992. Available at: https://catalog.ldc.upenn.edu/LDC97S62

[12] Budzianowski, P., Wen, T.-H., Tseng, B.-H., Casanueva, I., Ultes, S., Ramadan, O., & Gasic, M. (2018). MultiWOZ – A large-scale multi-domain Wizard-of-Oz dataset for task-oriented dialogue modelling. arXiv:1810.00278

[13] Hugging Face Hub Datasets Search for "Call center" https://huggingface.co/datasets?sort=trending&search=Call+center

[14] Gang Hu, Koren M. Jo, Yi Alex Wang, Jing Xie, Institutional trading and Abel Noser data, Journal of Corporate Finance, Volume 52, 2018, Pages 143-167, ISSN 0929-1199, https://doi.org/10.1016/j.jcorpfin.2018.08.005. (https://www.sciencedirect.com/science/article/pii/S0929119917307769).
17